\pdfoutput=1

\documentclass[11pt]{article}

\usepackage[final]{acl}

\usepackage{times}
\usepackage{latexsym}

\usepackage[T1]{fontenc}

\usepackage[utf8]{inputenc}

\usepackage{microtype}

\usepackage{inconsolata}

\usepackage{graphicx}

\usepackage{multirow}
\usepackage{graphicx}
\usepackage{comment}
\usepackage{amsmath, amsfonts}
\usepackage[linesnumbered,ruled,vlined]{algorithm2e}

\title{X-CoT: Explainable Text-to-Video Retrieval via LLM-based Chain-of-Thought Reasoning}

\author{Prasanna Reddy Pulakurthi$^{1}$,
Jiamian Wang$^{1}$,
Majid Rabbani$^{1}$,\\
{\bf Sohail Dianat$^{1}$},
{\bf Raghuveer Rao$^{2}$},
\and 
{\bf Zhiqiang Tao$^{1}$}\\
$^{1}$Rochester Institute of Technology, 
$^{2}$DEVCOM Army Research Laboratory
}

\begin{document}
\maketitle

\begin{abstract}
Prevalent text-to-video retrieval systems mainly adopt embedding models for feature extraction and compute cosine similarities for ranking. However, this design presents two limitations. Low-quality text-video data pairs could compromise the retrieval, yet are hard to identify and examine. Cosine similarity alone provides no explanation for the ranking results, limiting the interpretability. We ask that \textit{can we interpret the ranking results, so as to assess the retrieval models and examine the text-video data?} This work proposes X-CoT, an explainable retrieval framework upon LLM CoT reasoning in place of the embedding model-based similarity ranking. We first expand the existing benchmarks with additional video annotations to support semantic understanding and reduce data bias. We also devise a retrieval CoT consisting of pairwise comparison steps, yielding detailed reasoning and complete ranking. X-CoT empirically improves the retrieval performance and produces detailed rationales. It also facilitates the model behavior and data quality analysis. Code and data are available at: \href{https://github.com/PrasannaPulakurthi/X-CoT}{\textbf{\textit{github.com/PrasannaPulakurthi/X-CoT}}}.
\end{abstract}
    
\section{Introduction}
\label{sec:intro}


Text-to-video retrieval finds the most relevant video for a text query, being widely used for retrieval-augmented generation~\cite{jeong2025videorag}, question-answering~\cite{sun2024stllava}, and agent memory enhancement~\cite{fan2024embodied,sun2024elip}, etc.   
Recent progress mainly depends on embedding models, \emph{e.g.}, CLIP-based~\cite{ma2022x,wang2024text,wang2024diffusion} or MLLM-based~\cite{jiang2024vlm2vec,sun2024sq} for retrieval.




However, an embedding model-based retrieval system bears some limitations. First, the model is prone to the data quality of text-video pairs. Public datasets can introduce either flawed videos (\emph{e.g.}, blur, distortion) or crude captions~\cite{radford2021learning}, undermining the retrieval and making it hard to track. Second, the embedding model mainly computes the cosine similarity in the latent space, which only tells the ranking but fails to justify the ranking results. Both of these reasons call for an explainable retrieval system to interpret \textit{why a video candidate was retrieved}, so as to assist the users to comprehend the ranking results, assess the retrieval system, and examine the input data quality. 



\begin{figure}[t] 
\centering 
\includegraphics[width=0.95\columnwidth]{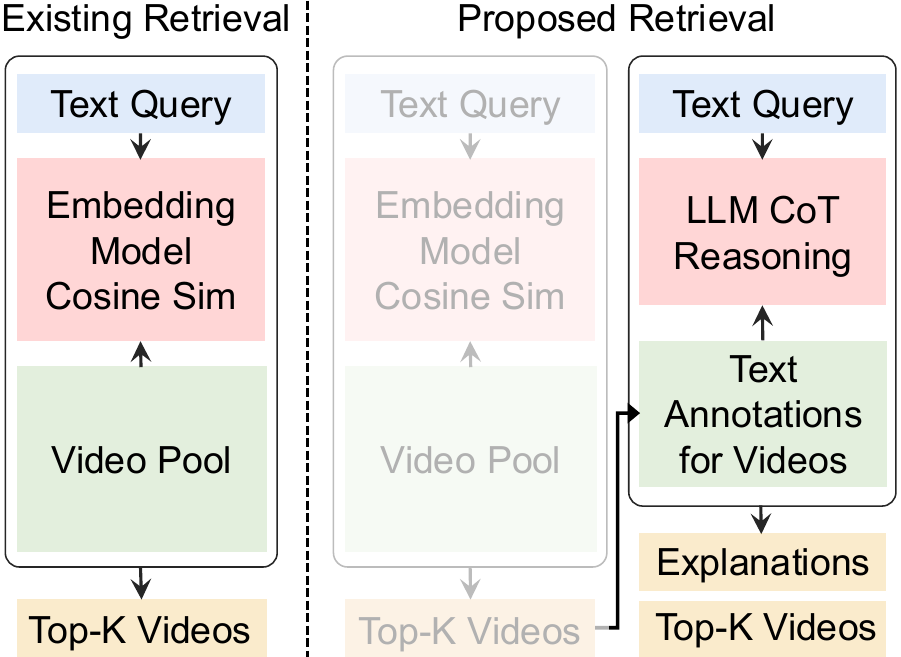}
\caption{
Existing retrieval systems mainly adopt embedding models to compute cosine similarities. We propose LLM CoT reasoning-based retrieval to provide explanations beyond rankings. Our method can also be integrated upon diverse embedding model methods. 
 }
\label{fig: intro}
\end{figure}


To achieve interpretability, this work proposes X-CoT, an explainable framework that exchanges traditional cosine similarity-based ranking with LLM-based judgment (see Fig.~\ref{fig: intro}) and devises a chain-of-thought pipeline for text-video retrieval. Firstly, we expand the existing benchmark datasets with additional video annotations to facilitate the LLM's reasoning and reduce the raw video data bias. Secondly, we define a retrieval CoT consisting of pairwise comparison steps upon the Bradley–Terry model~\cite{bradley1952rank}. By collecting the stepwise results, the proposed method not only enables the improved ranking performance over embedding model-based baselines but also delivers detailed rationales. In addition, without requiring the paired text-video data training, this method could serve as a general processing step that integrates with distinct embedding models. 

We summarize the contributions as follows:
(1) This work proposes X-CoT, an explainable retrieval system upon LLM chain-of-thought reasoning, advancing the trustworthy and trackable retrieval beyond the embedding model design. 
(2) We collect and release high-quality text annotation data for the raw videos to augment existing benchmark text-video datasets for future LLM study. 
(3) This work devises a retrieval CoT upon a pretrained LLM, being free of optimization and plug-and-play on top of the existing retrieval systems. 
(4) Experiments demonstrate the remarkable performance boost of X-CoT upon diverse embedding models and benchmark datasets. With X-CoT, we empirically analyze the behaviors of embedding models and identify the inferior text-video data.

\section{Related Work}
\label{sec: related work}

\textbf{Text-Video (T2V) Retrieval} has been driven by embedding models like X-CLIP~\cite{ma2022x}, Clip4clip~\cite{luo2021clip4clip}, Clip-vip~\cite{xue2022clip}, Cap4video~\cite{wu2023cap4video}, UMT~\cite{li2023unmasked}, and InternVid~\citep{wang2024internvid}, which learn joint video-text representations for retrieval.

\textbf{MLLMs for Retrieval.} Recent advances in MLLMs extend language models with visual understanding, enabling new capabilities in retrieval and reasoning. VLM2Vec~\cite{jiang2024vlm2vec} excels at text-image retrieval, having been trained for large-scale multimodal embedding tasks. MM-REACT~\cite{yang2023mm} combines visual tools with LLM reasoning. While Video-ChatGPT~\cite{maaz2023video} and Video-LLaVA~\cite{lin2023video} allow free-form video understanding through frame-by-frame perception and dialogue. BRIGHT~\cite{su2024bright} introduces a challenging benchmark focused on reasoning-intensive multimodal retrieval, highlighting the need for interpretable and robust systems like ours.

\section{Method}
\label{sec:method}

\subsection{Preliminaries}

Existing text-to-video retrieval systems are mainly embedding model-based.  
Given a video candidate $v$ and a text query $q$, an embedding model 
produces the video and text embedding, respectively, \emph{i.e.}, $\mathbf{z}_v,\mathbf{z}_q \in\mathbb{R}^d$, where $d$ denotes the dimension of the embedding space. Given the features, the system computes the cosine similarity score $s$ for ranking, \emph{i.e.},  $s(q,v)=(\mathbf{z}_q^{\top}\mathbf{z}_v) / (\lVert\mathbf{z}_q\rVert_2\lVert\mathbf{z}_v\rVert_2)$. However, it is hard to understand the rationale behind a specific cosine similarity score, \emph{e.g.}, \textit{what is the specific reason that the $s(q,v)$ is high/low for text $q$ and video $v$}, which could attribute to either text-video data correspondence or embedding models' behavior. To this end, this work studies explainable retrieval. 


\subsection{Video Annotation Collection}\label{subsec: data}


\textbf{Motivation.}
We first expand the existing text-video benchmarks with additional video annotations for the following reasons. 
(1) Videos can contain complex semantics, such as scenes with rapid motions or massive objects. Additional annotations provide a better chance for video understanding. 
(2) Video could be noisy and mislead the retrieval due to blur and distortion. Additional annotations provide useful information to describe the video semantics, reducing the bias caused by noisy frames. 




\begin{figure}[t] 
\centering 
\includegraphics[width=0.99\columnwidth]{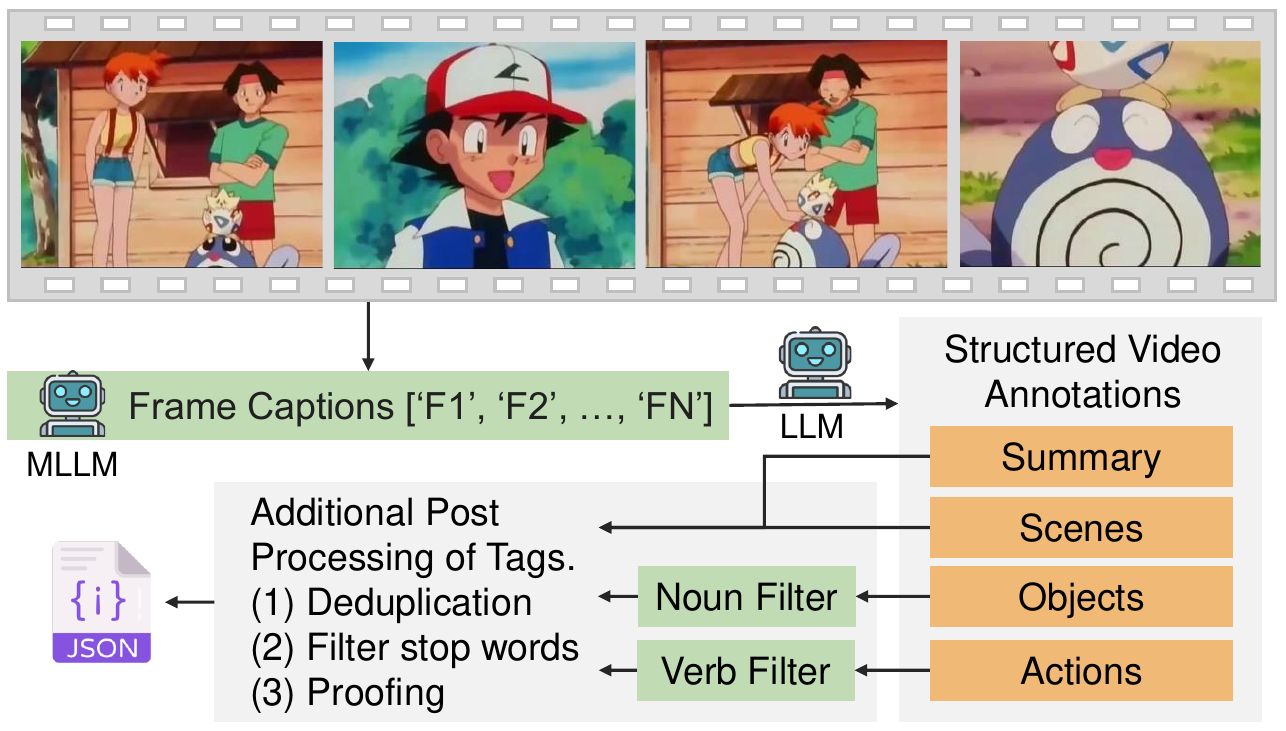}
\caption{Video annotation collection pipeline. Structured text is constructed to enrich the semantics and assist LLM reasoning. Ground-truth captions are not directly used.}
\label{fig: data_collection}
\end{figure}

\begin{figure}[t] 
\centering 
 \includegraphics[width=0.99\columnwidth]{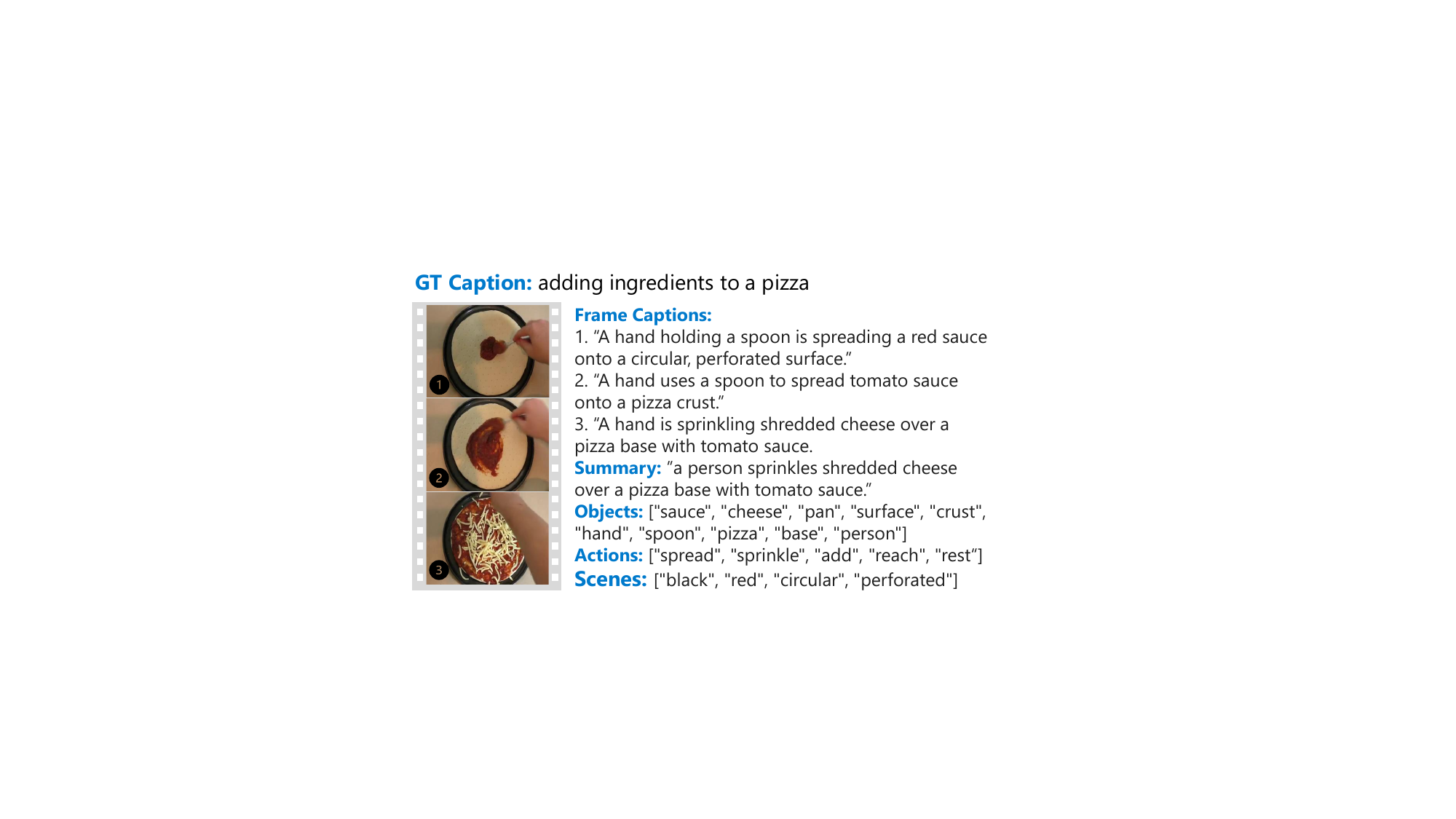}
\caption{Example of one structured video annotation.}
\label{fig: annotations}
\end{figure}

\textbf{Data Collection Pipeline.}
To collect the high-quality annotations, we develop an MLLM-based pipeline (see Fig.~\ref{fig: data_collection}). For every video $v$, we uniformly sample $N$ frames and apply the filters to remove near-duplicates (see Appendix~\ref{sec: frame_filtering}). We then adopt an MLLM (Qwen2.5-VL-7B-Captioner-Relaxed) to generate frame-level captions, which are aggregated and rephrased to form structured annotations comprising \texttt{objects}, \texttt{actions}, and \texttt{scenes}, plus a high-level video summary. 

We apply additional post-processing steps to improve annotation quality, including  
(i) \textbf{Noun Filter:} Extract and retain relevant object and scene tags for grounding entities.
(ii) \textbf{Verb Filter:} Extract action-related verbs to support temporal and causal reasoning.
(iii) \textbf{Deduplication:} Redundant or semantically equivalent tags (e.g., "a dog", "dog", "the dog") are merged to avoid repetition.
(iv) \textbf{Stop Word Removal:} Common stop words (e.g., "the", "is", "in") are filtered out to retain only informative content words.
(v) \textbf{Proofing:} Correct grammatical or formatting inconsistencies in the tags.
(vi) \textbf{Normalization:} We apply basic text normalization, including lowercasing and punctuation removal.
%
All videos are equipped with structured annotations, as illustrated in Fig.~\ref{fig: annotations}.




\subsection{Retrieval CoT}\label{subsec: X-CoT}

\begin{figure}[t] 
\centering 
\includegraphics[width=0.9\columnwidth]{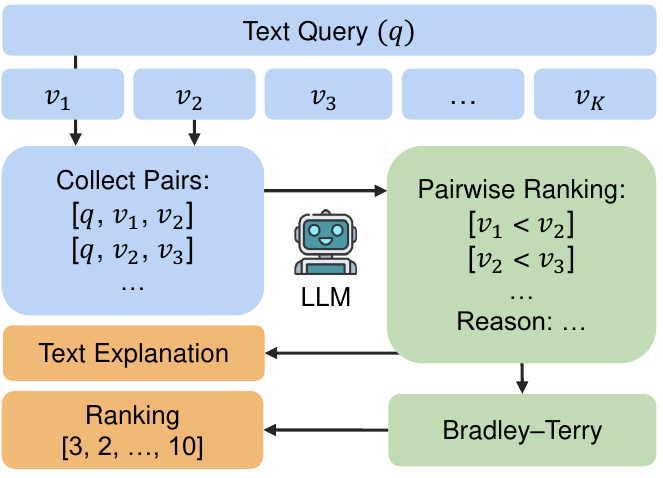}
\caption{X-CoT pipeline, which contains pairwise comparisons upon LLM for stepwise ranking and reasoning.}
\label{fig: pairwise_llm}
\end{figure}



Given the annotation data, this work adopts LLM reasoning for explainable retrieval. We construct a retrieval CoT to jointly produce the ranking and explanations, as shown in Fig.~\ref{fig: pairwise_llm}. The whole pipeline contains three steps. 

\textbf{Step 1}: One can optionally adopt diverse embedding models to produce top-$K$ candidate pool for a given query. Since the existing embedding model-based methods enable accurate retrieval with a large $K$ value, one can apply the proposed X-CoT to reason among a small range, \emph{e.g.}, $\mathcal{V}={\{v_1,\dots,v_K\}}$, $K<25$. 

\textbf{Step 2}:
We then generate pairwise combinations of the top‑$K$ candidates, forming input tuple $[q, v_i, v_j]$. We adopt LLM to process each tuple,  yielding the binary preference (\emph{e.g.}, $v_i < v_j$) and the text justification.
The structured annotations are employed to facilitate the reasoning.

\textbf{Step 3}: 
Notably, we further refine the ranking by approximating the Bradley–Terry (BT) model on the pairwise set via MLE~\cite{hunter2004mm} and compute the ability scores $\theta_k$ with $P_r[v_i>v_j]=\theta_i/(\theta_i+\theta_j)$. 
By this means, we correct the comparisons with noisy or cyclic judgments. Accordingly, the final ranking list $\hat{\mathcal{V}}$ is produced by 
Sorting in descending order. We provide the X-CoT algorithm in Appendix~\ref{sec: algo}. 


\section{Experiment}
\label{sec:experiment}

\subsection{Experimental Settings}\label{subsec: settings}

We evaluate X-CoT on four benchmarks: MSR-VTT~\cite{xu2016msr}, MSVD~\cite{chen2011collecting}, LSMDC~\cite{rohrbach2015dataset}, and DiDeMo~\cite{anne2017localizing}.
We report Recall@K (R@1, R@5, R@10), Median Rank (MdR), and Mean Rank (MnR). 

We consider three off-the-shelf embedding models to generate the coarse top-$K$ list ($K=20$), including CLIP-ViT-B/32~\cite{radford2021learning}, Qwen2-VL~\cite{wang2024qwen2} model by VLM2Vec~\cite{jiang2024vlm2vec},  and X-Pool~\cite{gorti2022x}. The former two are zero-shot retrievers, and X-Pool is trained with text-video data. 

\subsection{Performance Comparison}\label{subsec: performance}
Table~\ref{tab:benchmark_combined_msrvtt_msvd} and~\ref{tab: benchmark_combined_lsmdc_didemo} show the text-to-video retrieval performance with the proposed  X-CoT on four datasets and three embedding models. X-CoT enables a remarkable performance boost over embedding models on all metrics, \emph{e.g.}, $+5.6\%$ in R@1 for CLIP on MSVD, $+1.9\%$ in R@1 on MSVD for X-Pool. Overall, LLM CoT reasoning-based retrieval enjoys accurate retrieval over cosine similarity-based ranking upon embedding models.  



\begin{table*}[htb]
\begin{center}
\scalebox{0.77}{
\begin{tabular}{c|ccccc|ccccc}
\hline
\multirow{2}{*}{\textbf{Methods}} & \multicolumn{5}{c|}{\textbf{MSR-VTT}} & \multicolumn{5}{c}{\textbf{MSVD}} \\
\cline{2-11}
& \textbf{R@1}$\uparrow$ & \textbf{R@5}$\uparrow$ & \textbf{R@10}$\uparrow$ & \textbf{MdR}$\downarrow$ & \textbf{MnR}$\downarrow$ & \textbf{R@1}$\uparrow$ & \textbf{R@5}$\uparrow$ & \textbf{R@10}$\uparrow$ & \textbf{MdR}$\downarrow$ & \textbf{MnR}$\downarrow$ \\
\hline
How2Cap~\cite{shvetsova2024howtocaption} & 37.6 & 62.0 & 73.3 & 3.0 & --   & 44.5 & 73.3 & 82.1 & 2.0 & -- \\
TVTSv2~\cite{zeng2023tvtsv2} & 38.2 & 62.4 & 73.2 & 3.0 & --   & --   & --   & --   & --  & -- \\ 
InternVideo~\cite{wang2022internvideo} & 40.7 & 65.3 & 74.1 & 2.0 & --   & 43.4 & 69.9 & 79.1 & --   & -- \\ 
BT-Adapter~\cite{Liu_2024_CVPR} &  40.9 & 64.7 & 73.5 & -- & --   & -- & -- & -- & --   & -- \\ 
ViCLIP~\cite{wang2024internvid} & {42.4} & -- & -- & -- & --   & 49.1 & -- & -- & --   & -- \\ 
\hline
CLIP~\cite{radford2021learning} & 31.6 & 53.8 & 63.4 & 4.0 & 39.0 & 36.5 & 64.0 & 73.9 & 3.0 & 20.8 \\
X-CoT (ours) & \textbf{33.7} & \textbf{56.7} & \textbf{64.6} & \textbf{4.0} & \textbf{38.7} & \textbf{42.1} & \textbf{67.4} & \textbf{75.4} & \textbf{2.0} & \textbf{20.5} \\
\hline
VLM2Vec~\cite{jiang2024vlm2vec} & 36.4 & 60.2 & 70.7 & 3.0 & 27.3 & 46.7 & 73.8 & 82.6 & 2.0 & 12.8 \\
X-CoT (ours) & \textbf{37.2} & \textbf{61.8} & \textbf{71.5} & \textbf{3.0} & \textbf{27.1} & \textbf{48.4} & \textbf{74.8} & \textbf{83.2} & \textbf{2.0} & \textbf{12.6} \\
\hline
X-Pool~\citep{gorti2022x} & 46.9 & 73.0 & 82.0 & 2.0 & 14.2 & 47.2 & 77.2 & 86.0 & 2.0 & 9.3 \\
X-CoT (ours) & \textbf{47.3} & \textbf{73.3} & \textbf{82.1} & \textbf{2.0} & \textbf{14.2} & \textbf{49.1} & \textbf{78.0} & \textbf{86.6} & \textbf{2.0} & \textbf{9.2} \\
\hline
\end{tabular}}
\end{center}
\caption{Text-to-video retrieval performance comparison on MSR-VTT and MSVD.}
\label{tab:benchmark_combined_msrvtt_msvd}
\end{table*}

\begin{table*}[htb]
\begin{center}
\scalebox{0.77}{
\begin{tabular}{c|ccccc|ccccc} 
\hline 
\multirow{2}{*}{\textbf{Methods}} & \multicolumn{5}{c|}{\textbf{DiDeMo}} & \multicolumn{5}{c}{\textbf{LSMDC}} \\
\cline{2-11}
& \textbf{R@1}$\uparrow$ & \textbf{R@5}$\uparrow$ & \textbf{R@10}$\uparrow$ & \textbf{MdR}$\downarrow$  & \textbf{MnR}$\downarrow$ & \textbf{R@1}$\uparrow$ & \textbf{R@5}$\uparrow$ & \textbf{R@10}$\uparrow$ & \textbf{MdR}$\downarrow$  & \textbf{MnR}$\downarrow$ \\
\hline 
HiTeA~\cite{Ye_2023_ICCV} & 36.1 & 60.1 & 70.3 & --   & -- & 15.5 & 31.1 & 39.8 & --  & --    \\
TVTSv2~\cite{zeng2023tvtsv2}  & 34.6 & 61.9 & 71.5 & 3.0 & -- & 17.3 & 32.5 & 41.4 & 20.0 & --  \\ 
InternVideo~\cite{wang2022internvideo} & 31.5 & 57.6 & 68.2 & 3.0 & -- & 17.6 & 32.4 & 40.2 & 23.0 & --  \\
BT-Adapter~\cite{Liu_2024_CVPR} & 35.6 & 61.9 & 72.6 & --   & -- & 19.5 & 35.9 & 45.0 & -- & --    \\
ViCLIP~\cite{wang2024internvid} & 18.4 & -- & -- & --   & -- & 20.1 & -- & -- & -- & --    \\
\hline
CLIP~\cite{radford2021learning} & 25.2 & 49.4 & 59.0 & 6.0 & 49.7 & 15.9 & 28.4 & 35.3 & 31.0 & 129.6  \\
X-CoT (ours) &  \textbf{29.7} & \textbf{52.1} & \textbf{60.6} & \textbf{5.0} & \textbf{49.2} & \textbf{17.6} & \textbf{29.0} & \textbf{36.1} & \textbf{31.0} & \textbf{129.4}  \\
\hline
VLM2Vec~\cite{jiang2024vlm2vec}  & 33.5 & 57.7 & 68.4 & 4.0 & 34.1 & 18.2 & 33.6 & 41.4 & 23.0 & 119.1 \\
X-CoT (ours) & \textbf{35.8} & \textbf{59.2} & \textbf{68.8} & \textbf{3.0} & \textbf{33.9} & \textbf{18.9} & \textbf{35.1} & \textbf{41.9} & \textbf{23.0} & \textbf{118.9}  \\
\hline
X-Pool~\citep{gorti2022x} & 44.6 & 72.5 & 81.0 & 2.0 & 15.1 & 23.6 & 42.9 & 52.4 & 9.0 & 54.1  \\
X-CoT (ours) & \textbf{45.1} & \textbf{73.1} & \textbf{81.8} & \textbf{2.0} & \textbf{15.0} & \textbf{23.8} & \textbf{43.8} & \textbf{53.1} & \textbf{8.0} & \textbf{54.0}  \\
\hline
\end{tabular}}
\end{center}
\caption{Text-to-video retrieval performance comparison on DiDeMo and LSMDC.}
\label{tab: benchmark_combined_lsmdc_didemo}
\end{table*}

\subsection{Ablation Study}


\begin{table}[t]
\begin{center}
\scalebox{0.8}{
\begin{tabular}{c|ccccc} 
    \hline 
    \textbf{Method} & \textbf{R@1}$\uparrow$ & \textbf{R@5}$\uparrow$ & \textbf{R@10}$\uparrow$ & \textbf{MdR}$\downarrow$ & \textbf{MnR}$\downarrow$ \\
    \hline 
    Baseline & 25.2 & 49.4 & 59.0 & 6.0 & 49.7 \\
    w/o CoT & 22.3 & 39.4 & 58.9 & 6.0 & 49.7 \\
    w/o BT & 29.3 & 51.8 & 60.4 & 5.0 & 49.4\\
    \hline
    X-CoT & \textbf{29.7} & \textbf{52.1} & \textbf{60.6} & \textbf{5.0} & \textbf{49.2}\\
    \hline
\end{tabular}}
\end{center}
\caption{Ablation study of proposed X-CoT with CLIP-ViT-B/32 model ($K=20$) and upon DiDeMo Dataset.} 
\label{tab: cot_bt}
\end{table}

We conduct an ablation study toward the X-CoT in Table~\ref{tab: cot_bt}.  
We adopt the CLIP model as the baseline.
We study the effect of the proposed CoT with \texttt{w/o CoT}, \emph{i.e.}, directly ask the LLM to rank the top-K results, leading to a significant drop in performance, \emph{e.g.}, $-2.9\%$ for R@1 -- pairwise comparison is much easier than selecting best-of-K.  
We also find that the CoT model (\texttt{w/o BT}) benefits the retrieval. Jointly considering the CoT and the BT model, the proposed method improves the baseline by $4.5\%$ on R@1.



\begin{figure}[t] 
\centering 
\includegraphics[width=0.99\columnwidth]{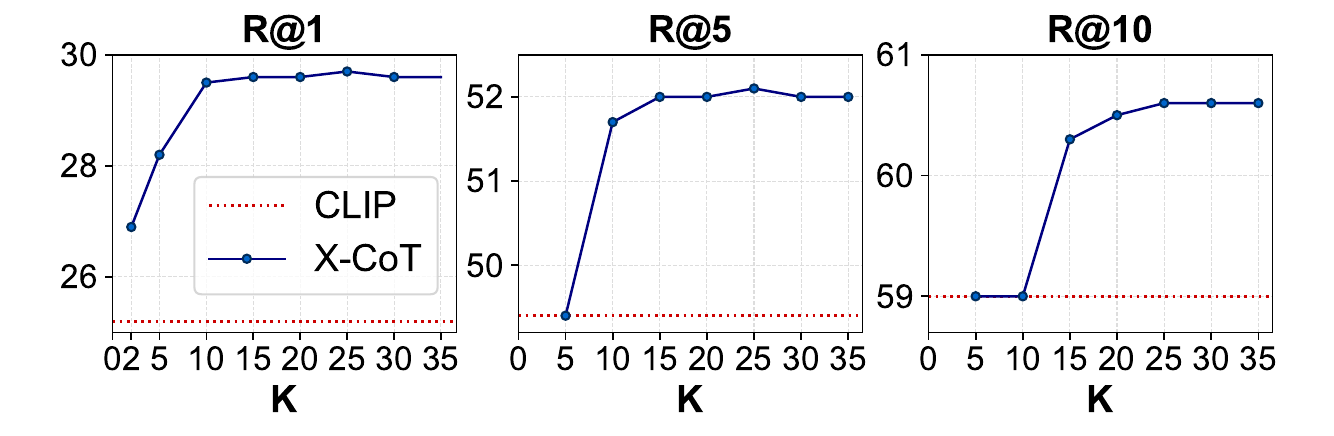}
\caption{top-$K$ discussion to facilitate X-CoT.  Performance reported with CLIP model on DiDeMo dataset.}
\label{fig: ablation_k}
\end{figure}

\subsection{Model Discussion}
\begin{figure}[t] 
\centering 
\includegraphics[width=0.99\columnwidth]{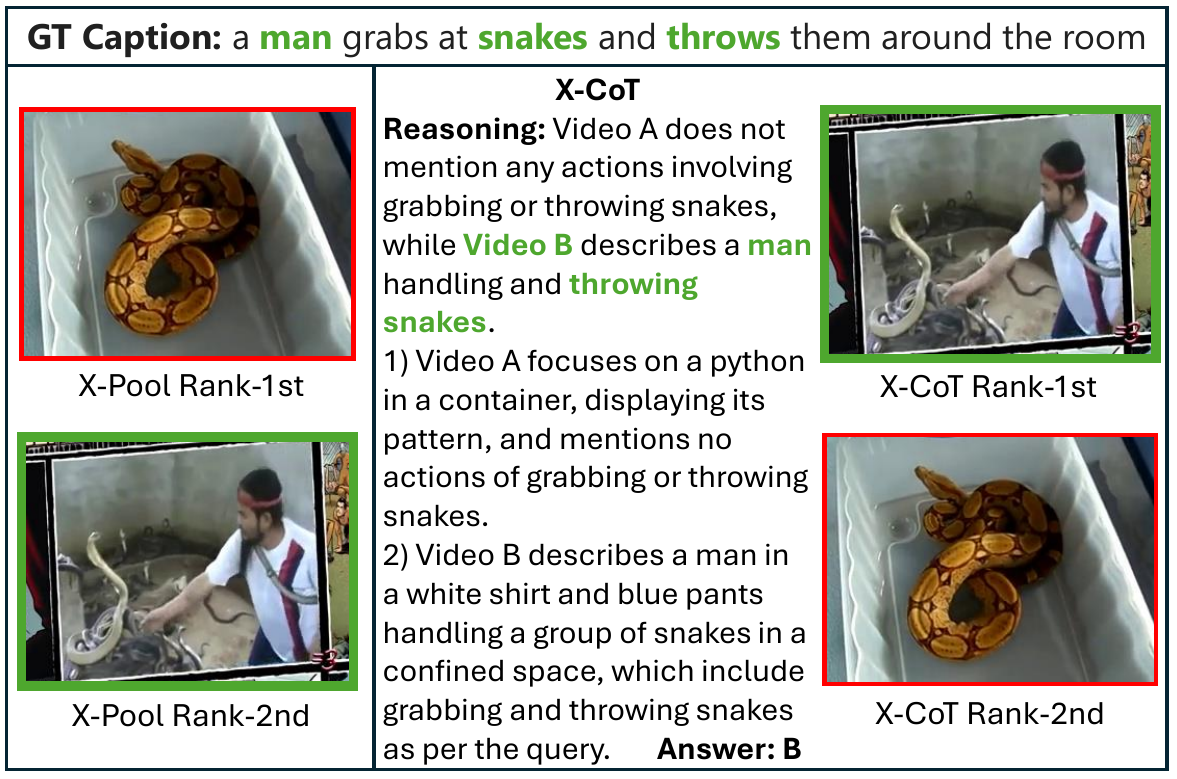}
\caption{Explainability discussion. X-Pool fails in ranking highly similar videos. By comparison, X-CoT identifies the relevant video, with subtle differences clearly explained.}
\label{fig: model_discussion_top2}
\end{figure}

\begin{figure}[thb] 
\centering 
\includegraphics[width=0.99\columnwidth]{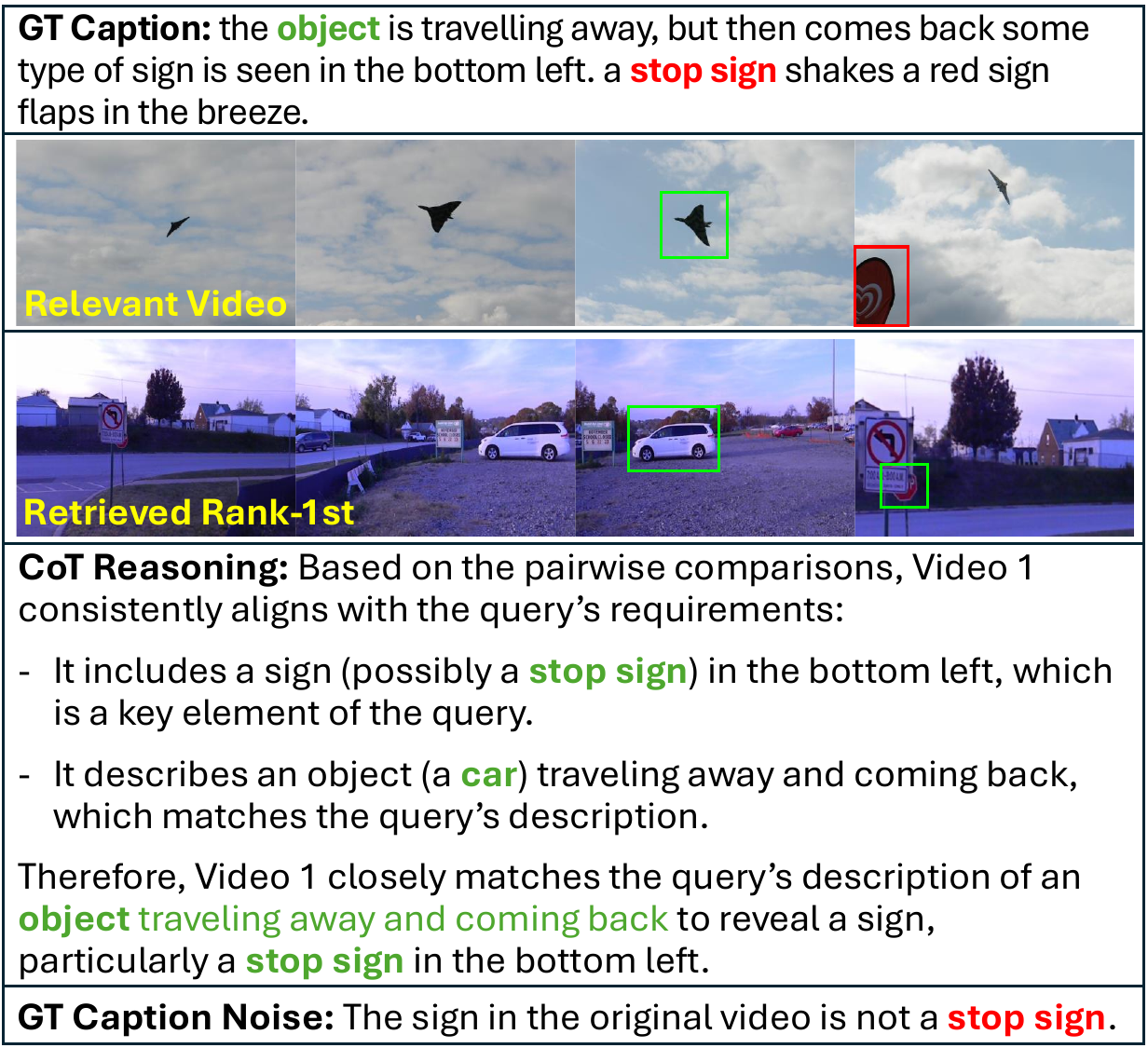}
\caption{Explainability discussion. By jointly examining the text caption, relevant video, and the CoT reasoning by X-CoT, one can find the ambiguous (\emph{e.g.}, object) and minor (\emph{e.g.}, stop sign) claims in the text caption, misleading the retrieval and introducing noise. 
}
\label{fig: gt_caption_errors}
\end{figure}

In Fig.~\ref{fig: ablation_k}, we discuss the top-$K$ ranges to facilitate X-CoT.  
X-CoT effectively identifies and ranks relevant candidates as $K$ grows, demonstrating an adaptivity to the pool scale. We further discuss the explainability of the proposed X-CoT. 
Fig.~\ref{fig: model_discussion_top2} discusses the explainability of X-CoT in evaluating the retrieval model's behavior. With explanations, one can diagnose the semantic factors that could be missed by the embedding model. 
\emph{e.g.}, the concept of ``man'' plays an important role. 
In addition, one can evaluate the text-video data quality with the proposed X-CoT. As shown in Fig.~\ref{fig: gt_caption_errors}, the proposed X-CoT fails for the given text query. However, the incorrect retrieval could be attributed to the text flaws by jointly examining the text caption, relevant video, and the CoT explanations. This demonstrates the power of the explainable retrieval system in the text-video data quality assessment. We provide success examples in Appendix~\ref{app: X-CoT_Ranking_Examples}.  



\section{Conclusion}
\label{sec:conclusion}
This work studied explainable retrieval systems and introduced X-CoT, an LLM CoT reasoning-based retrieval system in place of the embedding model cosine similarity-based ranking. To achieve the goal, we first expand the existing benchmarks with additional video annotation. We then constructed a pairwise CoT to provide reasoning and ranking. Experiments show X-CoT improves retrieval performance while providing explanations, demonstrating its potential for interpretable multimodal retrieval. We hope this work can inspire future endeavors in explainable retrieval. 


\section*{Limitations}
This work studies the explainable text-to-video retrieval upon LLM CoT reasoning. A potential limitation is that the reasoning and the ranking highly depend on the capacity of the LLM. 
While modern LLMs demonstrate strong generalization ability, they may be less effective in domain-specific or highly noisy text-video data scenarios, such as very long video comprehension. Considering that this could be one of the first efforts in this direction, we will explore more challenging text-to-video retrieval scenarios in future work. 

While the Bradley–Terry (BT) model provides a principled way to aggregate pairwise preferences, it also imposes certain constraints. The current formulation relies on binary win/loss outcomes and does not capture the uncertainty or nuanced reasoning strength that LLMs may provide. Future work could explore the incorporation of soft confidence scores or learnable aggregation strategies so that the richness of LLM reasoning in text-to-video retrieval can be better captured.

\bibliography{main}

\appendix

\clearpage
\appendix

\section{Similar Frame Filtering}
\label{sec: frame_filtering}


To ensure diversity in the frame annotations, we use a lightweight ResNet18~\cite{he2016deep} model pretrained on ImageNet~\cite{deng2009imagenet} to extract frame-level visual features. Each frame is resized, normalized, and passed through the network to obtain a feature embedding, which is L2-normalized. We then compare the current frame to all previously retained frames using cosine similarity, and if the maximum similarity is below a threshold (e.g., 0.95), the frame is kept. This process continues sequentially until the final set of non-duplicate frames is obtained, ensuring diversity and promoting frame-level annotation quality.

\section{Structured Video Annotations as Input: CLIP vs. X-CoT}
\label{app:clip_vs_xcot}
To test whether video annotations alone would suffice for CLIP, we use structured video annotations instead of the video embeddings and recompute cosine similarity with CLIP. As seen from Table~\ref{tab:app_clip_struct}, the performance drops compared to using X‑CoT, suggesting that LLM reasoning is required to exploit long, verb‑rich context.

\begin{table}[t]
\begin{center}
\scalebox{0.67}{
\begin{tabular}{l|ccccc}
\hline
\textbf{Methods} & \textbf{R@1}$\uparrow$ & \textbf{R@5}$\uparrow$ & \textbf{R@10}$\uparrow$ & \textbf{MdR}$\downarrow$ & \textbf{MnR}$\downarrow$ \\
\hline
Struct.\ ann.\ w/ CLIP & 16.9 & 30.5 & 39.1 & 25.5 & 141.9 \\
Struct.\ ann.\ w/ X-CoT & 33.7 & 56.7 & 64.6 & 4.0 & 38.7 \\
\hline
\end{tabular}}
\caption{Feeding structured video annotations to CLIP vs. using X‑CoT on the MSR‑VTT dataset.}
\label{tab:app_clip_struct}
\end{center}
\end{table}

\section{Robustness to Noisy Annotations}
\label{app:noisy}
To test the sensitivity of X-CoT to imperfect annotations, we perturb 20\% of tags in the structured annotations and re‑run X‑CoT on MSR‑VTT, as shown in Table~\ref{tab:app_noisy}. The proposed X-CoT experiences a small performance decline in the noisy scenario, demonstrating the robustness to the annotation data quality. We also observe that the complete annotation gives improved performance, showing the effectiveness of the collected annotation data.

\begin{table}[t]
\scalebox{0.67}{
\begin{tabular}{l|ccccc}
\hline
\textbf{Annotation Type} & \textbf{R@1}$\uparrow$ & \textbf{R@5}$\uparrow$ & \textbf{R@10}$\uparrow$ & \textbf{MdR}$\downarrow$ & \textbf{MnR}$\downarrow$ \\
\hline
20\% noisy tags & 32.3 & 53.9 & 62.0 & 4.0 & 49.1 \\
Complete annotations & 33.7 & 56.7 & 64.6 & 4.0 & 38.7 \\
\hline
\end{tabular}}
\caption{Effect of noisy structured annotations on X‑CoT (MSR‑VTT dataset).}
\label{tab:app_noisy}
\end{table}

\section{Additional Qualitative Video Annotation Examples}
\begin{figure}[t] 
\includegraphics[width=0.99\columnwidth]{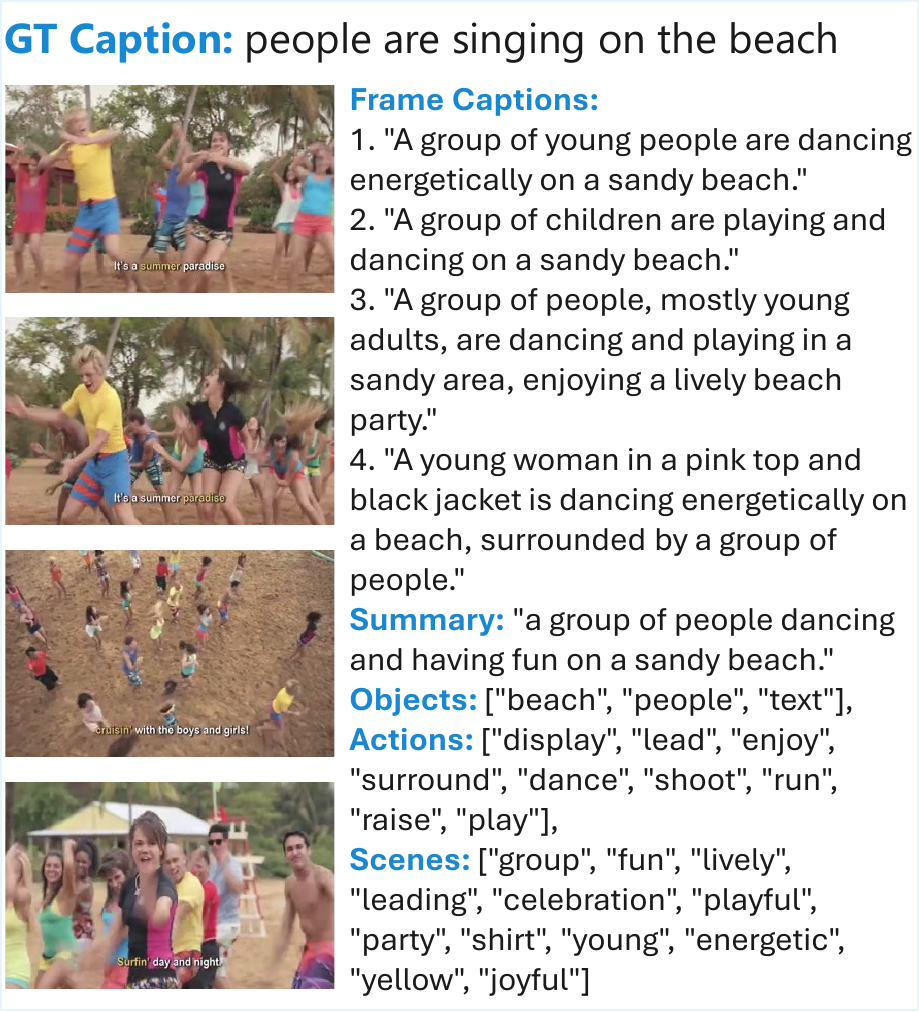}
\caption{Example of collected annotations.}
\label{fig: annotations_app1}
\end{figure}

\begin{figure}[t] 
\centering 
\includegraphics[width=0.99\columnwidth]{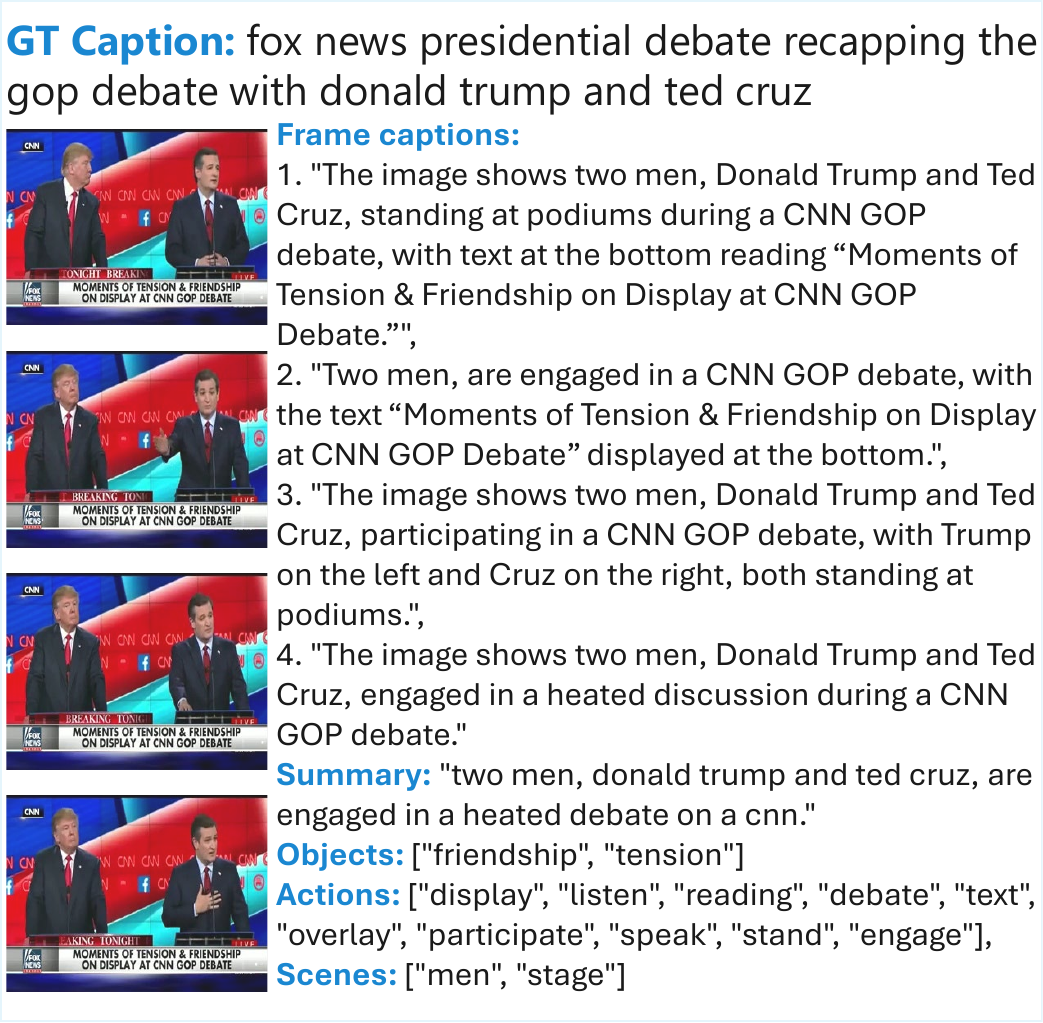}
\caption{Example of collected annotations.}
\label{fig: annotations_app2}
\end{figure}

Fig.~\ref{fig: annotations_app1} and Fig.~\ref{fig: annotations_app2} show examples where structured video annotations provide more accurate scene descriptions than the original dataset captions. These cases reveal:

\begin{enumerate}
    \item Semantic misalignment in GT labels as shown in Fig.~\ref{fig: annotations_app1} (e.g., labeling "dancing on a beach" as "singing").
    \item Fine-grained object and action detection as shown in Fig.~\ref{fig: annotations_app2} (e.g., political figures identified by name, or scene attributes like "joyful" or "heated").
\end{enumerate}

Such annotations serve as the foundation for X-CoT's reasoning mechanism and improve the overall retrieval reliability.

\section{Quantitative Evaluation of Video Annotations}

We introduce a proxy metric to assess the semantic faithfulness of the generated explanations. For each query in the MSR-VTT testing set, we record the top-1 video embedding $v_{\text{ori}}$ obtained from VLM2Vec. We then apply X-CoT to produce a re-ranked top-1 video embedding $v_{\text{xcot}}$ and the corresponding explanation embedding $e_{\text{expl}}$ (both derived from VLM2Vec). We compute the similarity between two types of video embedding as: 
\begin{align}
    sim_{\text{baseline}} & = \cos \langle e_{\text{expl}}, v_{\text{ori}} \rangle, \\ 
    sim_{\text{xcot}} & = \cos \langle e_{\text{expl}}, v_{\text{xcot}} \rangle .
\end{align}

Averaging these values across all queries yields $\bar{sim}_{\text{baseline}} = 0.273$ and $\bar{sim}_{\text{xcot}} = 0.350$. The $+0.077$ gain demonstrates that the explanation embeddings align more strongly with the X-CoT re-ranked results compared to the baseline retrieval, indicating that explanations are semantically faithful to the system’s final decision.  

To further guide future human-centered evaluation, established explanation-quality frameworks such as \cite{doshi2017towards} and \cite{deyoung-etal-2020-eraser} can be applied to assess interpretability and rationalization.

\section{X-CoT Pairwise Ranking Algorithm}\label{sec: algo}

The pseudo-code for the pairwise ranking is provided in Algorithm~\ref{alg:x-cot}.
Given the coarse \emph{top-$K$} list $V=[v_1,\dots ,v_K]$ (we set $K{=}20$),
X-CoT performs at most $P{=}10$ \emph{sliding-window} sweeps.
During each sweep, the list is scanned from left to right;
for every adjacent pair $(v_i,v_{i+1})$. An LLM receives the query plus two structured video descriptions
and must reply with its choice and reason.
If the answer favors $v_{i+1}$, the two items are swapped.

\textbf{Complexity.} In the best-case scenario, the number of pair-wise comparisons is $(K-1)$, and in the worst case, $P(K-1)$.

\textbf{LRU Caching.} The comparison routine is protected by an LRU cache keyed on the triple \mbox{$(\text{query},v_i,v_{i+1})$}.
Thus, although up to $(K-1)P=200$ comparisons are \emph{possible}, only $\sim$30-40 unique LLM calls are required on average,
saving $\approx85\%$ of LLM calls.

\textbf{Global Aggregation.} All newly observed win–loss edges are converted to ability scores $\theta_k$ via a Bradley–Terry maximum-likelihood fit (weak Gaussian prior $\alpha=10^{-3}$). Sorting $\theta_k$ in descending order yields the final ranking $\hat V$. In addition to the ranking, the individual explanations collected during each pairwise comparison are concatenated and summarized in a final single-shot LLM call.

\begin{algorithm*}[t]
\caption{\textsc{X-CoT Ranking via Pairwise Comparisons}}
\label{alg:x-cot}
\DontPrintSemicolon 
\SetKwComment{Comment}{\tcp}{}
\KwIn{%
Text query $q$;\\
Top-$K$ candidate list $\mathcal{V} = [v_1, \dots, v_K]$;\\
Number of passes $P = 10$;}
\KwOut{%
Sorted list $\hat{\mathcal{V}}$;\\
Pairwise explanation $\mathcal{R}$;\\
Final explanation $\mathcal{E}$;}

Initialize pairwise log $\mathcal{L} \leftarrow [\,]$ \tcp*{pairwise win log for Bradley–Terry} 
Initialize reason list $\mathcal{R} \leftarrow [\,]$ \tcp*{natural-language reasons from the LLM} 

\tcp{\textsc{CompareLLM}: takes query and a pair of candidates, returns the closed match to the query and a reason}
\tcp{\textsc{ExplainLLM}: summarizes the full set of pairwise reasons into a final explanation}

\For{$p \gets 1$ \KwTo $P$}{
  \For{$i \gets 1$ \KwTo $K-1$}{
    $(w, r) \leftarrow \textsc{CompareLLM}(q, \mathcal{V}[i], \mathcal{V}[i+1])$  \tcp*[r]{LLM returns winner $w$ and reason $r$}
    Append $r$ to $\mathcal{R}$, and $w$ to $\mathcal{L}$  \tcp*[r]{Log result and explanation}
    \If{$w = \mathcal{V}[i+1]$}{
      Swap $\mathcal{V}[i]$ and $\mathcal{V}[i+1]$ \tcp*[r]{If right candidate wins, swap positions}
    }
  }
}

$\hat{\mathcal{V}} \leftarrow \textsc{Bradley-Terry Aggregate}(\mathcal{L})$ \;
$\mathcal{E} \leftarrow \textsc{ExplainLLM}(\mathcal{R})$ \;

\Return $(\hat{\mathcal{V}}, \mathcal{E}, \mathcal{R})$ 
\end{algorithm*}

\section{Efficiency and Scalability}
\label{app:efficiency}

In Table~\ref{tab:app_efficiency}, we report the runtime and GPU memory cost under different hardware settings (e.g., number of NVIDIA RTX 3090 GPUs). As shown by Table~\ref{tab:app_efficiency}, the runtime per query could be drastically reduced as we scale the number of GPUs, being comparable with the CLIP-based embedding model (X-Pool) and the MLLM-based embedding model (VLM2Vec). This enhances the feasibility of real-world deployment. The above speedup is achieved by substantial engineering endeavors, including sliding window, caching, odd-even parallelization, and GPU parallelization.

\begin{table*}[t]
\begin{center}
\scalebox{0.84}{
\begin{tabular}{lccccccc}
\hline
\textbf{Methods (\#GPU)} & \textbf{X-CoT($\times$1)} & \textbf{X-CoT($\times$2)} & \textbf{X-CoT($\times$4)} & \textbf{X-CoT($\times$8)} & \textbf{X-CoT($\times$32)} & \textbf{X-Pool} & \textbf{VLM2Vec} \\
\hline
GPU Memory (GB) & 16.7 & 33.4 & 64.0 & 130.2 & 535.0 & 4.0 & 16.6 \\
Runtime / query (s) & 3.6 & 1.8 & 0.9 & 0.45 & 0.10 & 0.11 & 0.88 \\
\hline
\end{tabular}}
\caption{Runtime and memory profile of X‑CoT with increasing GPU parallelism alongside embedding‑based retrieval baselines. A local open‑source LLM (Qwen~2.5‑7B‑Instruct‑1M) was used (no API cost).}
\label{tab:app_efficiency}
\end{center}
\end{table*}

\textbf{Sliding Window and Caching.} Since the embedding model already provides a good initial ranking, our proposed method, which builds atop embedding models, only needs to perform a small number of local swaps, rather than running a total of $K(K-1) = 380$ LLM calls for top-20 ($K=20$) candidates per query. We adopt a sliding window strategy that compares only adjacent video pairs (e.g., (v1, v2), (v2, v3), ..., ) across multiple passes. Since many of the pairwise comparisons recur across the passes, we cache the pairwise results to avoid repetitive LLM calls. We empirically find that such a strategy can reduce the total number of LLM calls per query by 90\% on average (e.g., less than 40 LLM calls per query).

\textbf{Odd-Even Parallelization.} In each sliding window pass, for $K=20$ there will be 19 adjacent pairs. We partition these pairs into odd (e.g., (v1, v2), (v3, v4), ..., (v19, v20)) and even (e.g., (v2, v3), (v4, v5), ..., (v18, v19)) groups, where both the odd and even groups consist of non-overlapping pairs. The comparisons within each group are executed in parallel via multi-threaded dispatch, thereby reducing the wall-clock latency of each pass.

\textbf{GPU Parallelization.} For each query, multiple LLM calls (i.e., pairwise comparisons) are independent and can be parallelized. We leverage GPU-level concurrency to distribute the LLM calls across multiple devices. Together with the above engineering strategies, we reduce the latency as shown in Table~\ref{tab:app_efficiency}.

Since we adopt the open-source LLM (Qwen 2.5-7B-Instruct-1M) and the local hardware, no direct monetary cost is incurred.

\section{X-CoT Ranking Examples}
\label{app: X-CoT_Ranking_Examples}
\begin{figure*}[t] 
\centering 
\includegraphics[width=1.99\columnwidth]{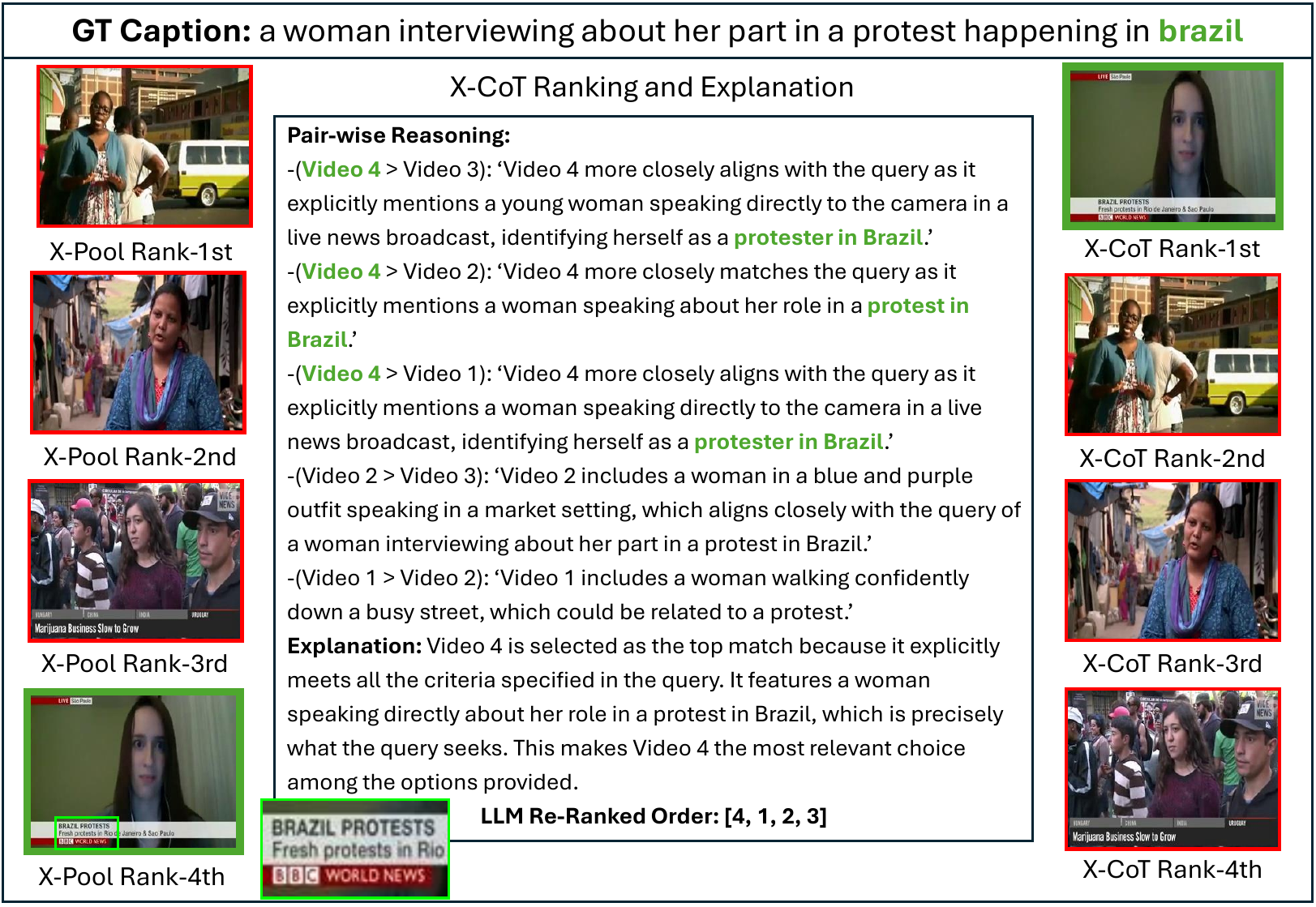}
\caption{Successful ranking with X-CoT on a query about a protest in Brazil. The top result is selected through stepwise pairwise comparisons, supported by natural language justifications.}
\label{fig: model_discussion_top4}
\end{figure*}

Fig.~\ref{fig: model_discussion_top4} illustrates how our method re-ranks candidate videos through pairwise reasoning and global aggregation. From the multiple pairwise judgments, culminating in the accurate re-ranking of a video showing a protester in Brazil speaking to a reporter, precisely matching the query.

\section{Embedding Model Details and Complete Benchmarking Results}
We evaluate two zero-shot models, CLIP~\cite{radford2021learning} and VLM2Vec~\cite{jiang2024vlm2vec}, alongside a fine-tuned model, X-Pool~\cite{gorti2022x}, to assess retrieval performance across diverse settings. The complete benchmarking results for MSR-VTT~\cite{xu2016msr} and MSVD~\cite{chen2011collecting} are presented in Table~\ref{tab:benchmark_combined_msrvtt_msvd_appendix}, and for DiDeMo~\cite{anne2017localizing} and LSMDC~\cite{rohrbach2015dataset} are presented in Table~\ref{tab: benchmark_combined_lsmdc_didemo_appendix}.

\begin{table*}[t]
\begin{center}
\scalebox{0.77}{
\begin{tabular}{c|ccccc|ccccc}
\hline
\multirow{2}{*}{\textbf{Methods}} & \multicolumn{5}{c|}{\textbf{MSR-VTT}} & \multicolumn{5}{c}{\textbf{MSVD}} \\
\cline{2-11}
& \textbf{R@1}$\uparrow$ & \textbf{R@5}$\uparrow$ & \textbf{R@10}$\uparrow$ & \textbf{MdR}$\downarrow$ & \textbf{MnR}$\downarrow$ & \textbf{R@1}$\uparrow$ & \textbf{R@5}$\uparrow$ & \textbf{R@10}$\uparrow$ & \textbf{MdR}$\downarrow$ & \textbf{MnR}$\downarrow$ \\
\hline
ALPRO~\cite{li2022align} & 24.1 & 44.7 & 55.4 & 8.0  & --   & -- & -- & --   & --  & -- \\
BridgeFormer~\cite{Ge_2022_CVPR} & 26.0 & 46.4 & 56.4 & 7.0  & --   & 43.6 & 74.9  & 84.9   & 2.0  & -- \\ 
MILES~\cite{ge2022miles} & 26.1 & 47.2 & 56.9 & 7.0  & --   & 44.4 & 76.2  & \textbf{87.0}   & 2.0  & -- \\
HiTeA~\cite{Ye_2023_ICCV} & 29.9 & 54.2 & 62.9 & --  & --   & -- & -- & -- & --   & -- \\
OmniVL~\cite{wang2022omnivl} & 34.6 & 58.4 & 66.6 & -- & --   & -- & -- & -- & -- & -- \\
ImageBind~\cite{girdhar2023imagebind} & 36.8 & 61.8 & 70.0 & --  & --   & --   & --   & --   & --  & -- \\
How2Cap~\cite{shvetsova2024howtocaption} & 37.6 & 62.0 & 73.3 & 3.0 & --   & 44.5 & 73.3 & 82.1 & 2.0 & -- \\
TVTSv2~\cite{zeng2023tvtsv2} & 38.2 & 62.4 & 73.2 & 3.0 & --   & --   & --   & --   & --  & -- \\ 
InternVideo~\cite{wang2022internvideo} & 40.7 & 65.3 & 74.1 & 2.0 & --   & 43.4 & 69.9 & 79.1 & --   & -- \\ 
BT-Adapter~\cite{Liu_2024_CVPR} &  40.9 & 64.7 & 73.5 & -- & --   & -- & -- & -- & --   & -- \\ 
ViCLIP~\cite{wang2024internvid} & {42.4} & -- & -- & -- & --   & 49.1 & -- & -- & --   & -- \\ 
LanguageBind~\cite{zhulanguagebind}  & 42.6 & 65.4 & 75.5 & --  & --   & 52.2 & 79.4 & 87.3 & --   & -- \\
LamRA~\cite{liu2024lamra} & 44.7 & 68.6 & 78.6 & --  & --   & 52.4 & 79.8 & 87.0 & --   & -- \\
\hline
CLIP~\cite{radford2021learning} & 31.6 & 53.8 & 63.4 & 4.0 & 39.0 & 36.5 & 64.0 & 73.9 & 3.0 & 20.8 \\
X-CoT (ours) & \textbf{33.7} & \textbf{56.7} & \textbf{64.6} & \textbf{4.0} & \textbf{38.7} & \textbf{42.1} & \textbf{67.4} & \textbf{75.4} & \textbf{2.0} & \textbf{20.5} \\
\hline
VLM2Vec~\cite{jiang2024vlm2vec} & 36.4 & 60.2 & 70.7 & 3.0 & 27.3 & 46.7 & 73.8 & 82.6 & 2.0 & 12.8 \\
X-CoT (ours) & \textbf{37.2} & \textbf{61.8} & \textbf{71.5} & \textbf{3.0} & \textbf{27.1} & \textbf{48.4} & \textbf{74.8} & \textbf{83.2} & \textbf{2.0} & \textbf{12.6} \\
\hline
X-Pool~\citep{gorti2022x} & 46.9 & 73.0 & 82.0 & 2.0 & 14.2 & 47.2 & 77.2 & 86.0 & 2.0 & 9.3 \\
X-CoT (ours) & \textbf{47.3} & \textbf{73.3} & \textbf{82.1} & \textbf{2.0} & \textbf{14.2} & \textbf{49.1} & \textbf{78.0} & \textbf{86.6} & \textbf{2.0} & \textbf{9.2} \\
\hline
\end{tabular}}
\end{center}
\caption{Complete Text-to-video retrieval performance comparison on MSR-VTT and MSVD.}
\label{tab:benchmark_combined_msrvtt_msvd_appendix}
\end{table*}

\begin{table*}[t]
\begin{center}
\scalebox{0.77}{
\begin{tabular}{c|ccccc|ccccc} 
\hline 
\multirow{2}{*}{\textbf{Methods}} & \multicolumn{5}{c|}{\textbf{DiDeMo}} & \multicolumn{5}{c}{\textbf{LSMDC}} \\
\cline{2-11}
& \textbf{R@1}$\uparrow$ & \textbf{R@5}$\uparrow$ & \textbf{R@10}$\uparrow$ & \textbf{MdR}$\downarrow$  & \textbf{MnR}$\downarrow$ & \textbf{R@1}$\uparrow$ & \textbf{R@5}$\uparrow$ & \textbf{R@10}$\uparrow$ & \textbf{MdR}$\downarrow$  & \textbf{MnR}$\downarrow$ \\
\hline 
ALPRO~\cite{li2022align}  & 23.8 & 47.3 & 57.9 & 6.0 & -- & --   & -- & -- & --   & --   \\
BridgeFormer~\cite{Ge_2022_CVPR} & 25.6 & 50.6  & 61.1   & 5.0  & -- & 12.2 & 25.9 & 32.2 & 42.0  & --    \\
MILES~\cite{ge2022miles} & 27.2 & 50.3  & 63.6   & 5.0  & -- & 11.1 & 24.7 & 30.6 & 50.7  & --    \\
HiTeA~\cite{Ye_2023_ICCV} & 36.1 & 60.1 & 70.3 & --   & -- & 15.5 & 31.1 & 39.8 & --  & --    \\
OmniVL~\cite{wang2022omnivl} & 33.3 & 58.7   & 68.5 & -- & -- & -- & -- & -- & -- & --  \\
How2Cap~\cite{shvetsova2024howtocaption}  & -- & -- & -- & -- & -- & -- &  17.3 & 31.7 & 38.6 & 29.0  \\
TVTSv2~\cite{zeng2023tvtsv2}  & 34.6 & 61.9 & 71.5 & 3.0 & -- & 17.3 & 32.5 & 41.4 & 20.0 & --  \\ 
InternVideo~\cite{wang2022internvideo} & 31.5 & 57.6 & 68.2 & 3.0 & -- & 17.6 & 32.4 & 40.2 & 23.0 & --  \\
BT-Adapter~\cite{Liu_2024_CVPR} & 35.6 & 61.9 & 72.6 & --   & -- & 19.5 & 35.9 & 45.0 & -- & --    \\
ViCLIP~\cite{wang2024internvid} & 18.4 & -- & -- & --   & -- & 20.1 & -- & -- & -- & --    \\
LanguageBind~\cite{zhulanguagebind} & 37.8 & 63.2 & 73.4 & -- & --  & -- &  -- & -- & -- & --  \\ 
\hline
CLIP~\cite{radford2021learning} & 25.2 & 49.4 & 59.0 & 6.0 & 49.7 & 15.9 & 28.4 & 35.3 & 31.0 & 129.6  \\
X-CoT (ours) &  \textbf{29.7} & \textbf{52.1} & \textbf{60.6} & \textbf{5.0} & \textbf{49.2} & \textbf{17.6} & \textbf{29.0} & \textbf{36.1} & \textbf{31.0} & \textbf{129.4}  \\
\hline
VLM2Vec~\cite{jiang2024vlm2vec}  & 33.5 & 57.7 & 68.4 & 4.0 & 34.1 & 18.2 & 33.6 & 41.4 & 23.0 & 119.1 \\
X-CoT (ours) & \textbf{35.8} & \textbf{59.2} & \textbf{68.8} & \textbf{3.0} & \textbf{33.9} & \textbf{18.9} & \textbf{35.1} & \textbf{41.9} & \textbf{23.0} & \textbf{118.9}  \\
\hline
X-Pool~\citep{gorti2022x} & 44.6 & 72.5 & 81.0 & 2.0 & 15.1 & 23.6 & 42.9 & 52.4 & 9.0 & 54.1  \\
X-CoT (ours) & \textbf{45.1} & \textbf{73.1} & \textbf{81.8} & \textbf{2.0} & \textbf{15.0} & \textbf{23.8} & \textbf{43.8} & \textbf{53.1} & \textbf{8.0} & \textbf{54.0}  \\
\hline
\end{tabular}}
\end{center}
\caption{Complete Text-to-video retrieval performance comparison on DiDeMo and LSMDC.}
\label{tab: benchmark_combined_lsmdc_didemo_appendix}
\end{table*}

\end{document}